\documentclass{llncs}
\usepackage{times} \usepackage{graphicx}
\usepackage{latexsym} \usepackage{amsmath, amssymb}
\usepackage{url}
\begin{document}
\title{Earthquake Scenario Reduction\\
by Symmetry Reasoning}
\author{Steven Prestwich}
\institute{Cork Constraint Computation Centre, Dept of Computer Science, \\
 University College, Cork, Ireland \\
\url{s.prestwich@cs.ucc.ie}
}

\maketitle
\begin{abstract}

A recently identified problem is that of finding an optimal investment
plan for a transportation network, given that a disaster such as an
earthquake may destroy links in the network.  The aim is to strengthen
key links to preserve the expected network connectivity.  A network
based on the Istanbul highway system has thirty links and therefore a
billion scenarios, but it has been estimated that sampling a million
scenarios gives reasonable accuracy.  In this paper we use symmetry
reasoning to reduce the number of scenarios to a much smaller number,
making sampling unnecessary.  This result can be used to facilitate
metaheuristic and exact approaches to the problem.

\end{abstract}

\section{Introduction}

We consider a known problem in pre-disaster planning: forming an
investment plan for a transportation network, with the aim of
facilitating rescue operations in the case of earthquakes.
Multi-stage stochastic problems occur in many real-life situations and
are often tackled by Stochastic Programming (SP) methods based on
Integer or Mathematical Programming.  These methods are guaranteed to
find an optimal solutions, but because of the complexity of the
problems they may only be practicable for small instances.

The use of metaheuristics such as Tabu Search, Simulated Annealing,
Genetic Algorithms and Ant Colony Optimisation is another promising
approach to such problems.  Though not guaranteed to find optimal
solutions, metaheuristics can often find near-optimal solutions in a
reasonable time.  But applying metaheuristics requires the computation
of an objective function (or fitness), which can be prohibitively
expensive for stochastic problems.  A common way of reducing the
computational effort is approximation by scenario sampling.

We propose an alternative method to sampling for the earthquake
problem, which does not involve approximation but which makes the
fitness computation tractable.  The paper is organised as follows.
Section \ref{problem} describes the problem and an existing approach
to solving it.  Section \ref{heur} outlines a standard metaheuristic
approach and points out its impracticality.  Section \ref{method}
describes our method.  Section \ref{conclusion} concludes the paper.

\section{A disaster pre-planning problem} \label{problem}

\begin{figure}
\begin{center}
\includegraphics[scale=0.5]{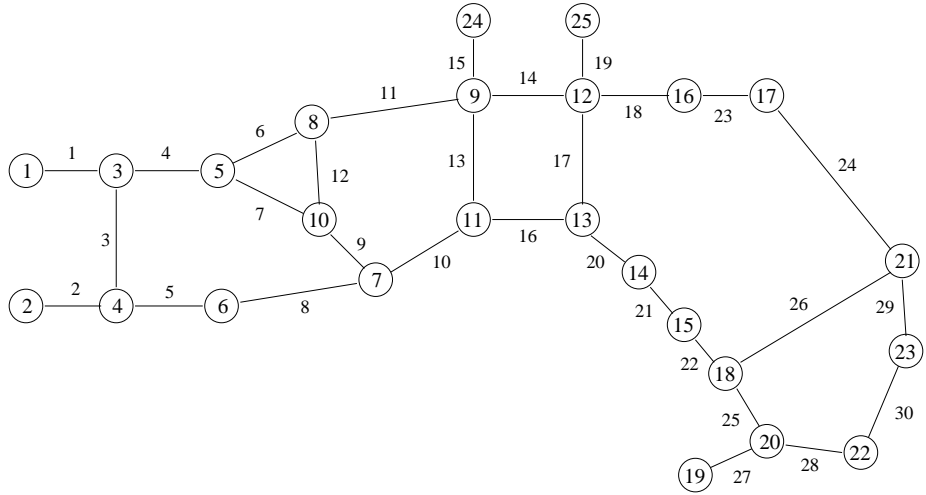}
\end{center}
\caption{Istanbul road network}
\label{istanbul}
\end{figure}

The problem is taken from Peeta {\it et al.\/} \cite{PeeEtc}.
Consider the transportation network in Figure \ref{istanbul} (not
drawn to scale), with nodes numbered 1--25 and arcs numbered 1--30,
each of whose arcs (which we shall refer to as {\it links\/}) may fail
with some probability.  This network is modelled on the Istanbul
highway network.

The failure probability of a link can be reduced by investing money in
it, and we have a budget limiting the total investment.  We aim to
minimise the expected shortest path between a specified source and
sink node in the network.  More generally, we aim to minimise a
weighted sum of shortest path lengths between several source-sink
pairs, chosen (for example) to represent paths between hospitals and
areas of high population.  This is an example of pre-disaster
planning, where a decision maker aims to maximise the robustness of a
transportation network with respect to possible disasters in order to
facilitate rescue operations.

First some notation.  Represent the network as an undirected graph
$G=(V,E)$ with nodes $V$ and arcs or links $E$.  For each link $e \in
E$ define a binary decision variable $y_e$ which is 1 if we invest in
that link and 0 otherwise.  Define a binary random variable $r_e$
which is 1 if link $e$ survives and 0 if it fails.  Denote the
survival (non-failure) probability of link $e$ by $p_e$ without
investment and $q_e$ with, the investment required for link $e$ by
$c_e$, the length of link $e$ by $t_e$, and the budget by $B$.  If
source and sink are unconnected then the path length is taken to be a
fixed number $M$ representing (for example) the cost of using a
helicopter.  Actually, if they are only connected by long paths then
they are considered to be unconnected, as in practice rescuers would
resort to alternatives such as rescue by helicopter or sea.  So Peeta
{\it et al.\/} only consider a few shortest paths for each source-sink
pair, as shown in Table \ref{shortpaths}.  We shall refer to these as
the {\it allowed paths\/}.  In each case $M$ is chosen to be the
smallest integer that is greater than the longest allowed path length.
They also consider a larger value of $M=120$ that places a greater
importance on connectivity.  Let us replace $M$ by 2 new constants:
$M_a$ is the length below which a path is allowed, while $M_p$ is the
penalty imposed when no allowed path exists.

\begin{table}
\begin{center}
\begin{tabular}{|r|rrrrrrrr|r|r|}
\hline
pair & \multicolumn{8}{|c|}{links} & length & $M_a$\\
\hline
14--20
& 21 & 22 & 25 & & & & & & 6.65 & \\
& 21 & 22 & 26 & 29 & 30 & 28 &  &  & 20.41 & \\
& 20 & 17 & 18 & 23 & 24 & 26 & 25 &  & 29.20 & \\
& 20 & 17 & 18 & 23 & 24 & 29 & 30 & 28 & 30.27 & 31\\
\hline
14--7
& 20 & 16 & 10 &  &  &  &  &  & 11.14 & \\
& 20 & 17 & 14 & 13 & 10 &  &  &  & 20.09 & \\
& 20 & 17 & 14 & 11 & 12 & 9 &  &  & 25.48 & \\
& 20 & 16 & 13 & 11 & 12 & 9 &  &  & 26.58 & \\
& 20 & 17 & 14 & 11 & 6 & 7 & 9 &  & 29.08 & \\
& 20 & 16 & 13 & 11 & 6 & 7 & 9 &  & 30.17 & 31\\
\hline
12--18
& 17 & 20 & 21 & 22 &  &  &  &  & 9.86 & \\
& 14 & 13 & 16 & 20 & 21 & 22 &  &  & 20.05 & \\
& 18 & 23 & 24 & 26 &  &  &  &  & 20.24 & \\
& 18 & 23 & 24 & 29 & 30 & 28 & 25 &  & 27.06 & 28\\
\hline
9--7
& 13 & 10 &  &  &  &  &  &  & 9.46 & \\
& 11 & 12 & 9 &  &  &  &  &  & 14.85 & \\
& 14 & 17 & 16 & 10 &  &  &  &  & 16.88 & \\
& 11 & 6 & 7 & 0 &  &  &  &  & 18.45 & 19\\
\hline
4--8
& 3 & 4 & 6 &  &  &  &  &  & 14.00 & \\
& 5 & 8 & 9 & 12 &  &  &  &  & 17.91 & \\
& 3 & 4 & 7 & 12 &  &  &  &  & 18.79 & \\
& 5 & 8 & 9 & 7 & 6 &  &  &  & 21.51 & \\
& 5 & 8 & 10 & 13 & 11 &  &  &  & 26.73 & \\
& 5 & 8 & 10 & 16 & 17 & 14 & 11 &  & 34.15 & 35\\
\hline
\end{tabular}
\end{center}
\caption{Allowed paths}
\label{shortpaths}
\end{table}

In SP terms this is a 2-stage problem.  In the first stage we must
decide which links to invest in, then link failures occur randomly.
In the second stage we must choose a shortest path between the source
and sink (the recourse action), given the surviving links.  If the
source and sink are no longer connected by an allowed path then a
fixed penalty $M_p$ is imposed.  Peeta {\it et al.\/} point out that,
though a natural approach is to strengthen the weakest links, this
does not necessarily lead to the best results.

This is a challenging problem because each of the 30 links might
independently be affected by earthquakes, giving $2^{30}$ scenarios.
Though optimisation time is not critical in pre-disaster planning,
over 1 billion scenarios is too many to be tractable.  Another source
of difficulty is that the problem has {\it endogenous uncertainty\/}:
the decisions (which links to invest in) affects the probabilities of
the random events (the link failures).  Relatively little work has
been done on such problems but they are usually much harder to solve
by SP methods.  For a survey on problems with endogenous uncertainty
see \cite{GoeGro}, which mentions applications including network
design and interdiction, server selection, facility location, and gas
reservoir development.  Other examples include clinical trial planning
\cite{ColMar} and portfolio optimisation \cite{SolEtc}.

\section{A metaheuristic approach} \label{heur}

Peeta {\it et al.\/} sample a million scenarios, and approximate the
objective function by a monotonic multilinear function.  They show
that their method gives optimal or near-optimal results on smaller
instances.  We are interested in applying standard metaheuristics such
as genetic algorithms to this problem.  As noted in a recent survey of
metaheuristic approaches to stochastic problems \cite{BiaEtc}, most
research is on continuous problems and problems with {\it noisy\/} or
{\it time-varying\/} fitness, and less work has been done on
metaheuristics for multi-stage problems.

An obvious approach to the earthquake problem is to use a population
of chromosomes, each with 30 binary genes corresponding to $y_1 \ldots
y_{30}$.  Thus each chromosome is a direct representation of an
investment plan in which values of 1 indicate investment and 0 no
investment.  Standard genetic operators (selection, recombination and
mutation) can be applied to this model.  To compute the fitness
(objective function) of a chromosome we check every scenario, each
representing a network realisation.  For a given network realisation
we compute the length of shortest paths between source-sink pairs
using (for example) Dijkstra's algorithm, taking value $M_p$ if there
is no path.  From all the scenarios we can compute expected path
lengths and hence the chromosome fitness.

A slight complication is the budget constraint: a chromosome might
contain too many 1-values, corresponding to overspend.  There are 3
ways of handling constraints in genetic algorithms \cite{CraEtc}:
\begin{itemize}
\item
Penalise constraint violation by adding a {\it penalty function\/} to
the fitness.
\item
{\it Repair\/} the chromosome so that it no longer violates any
constraints.
\item
Use a {\it decoder\/} to generate a feasible solution from the
chromosome, which is treated not as a solution but as a set of
instructions on how to construct one.
\end{itemize}
Any of these approaches can be applied to this problem, and we propose
a simple decoder: consider the genes in a fixed order (the numerical
order of links 1--30) and treat any 1-value that would violate the
budget constraint as a 0-value.  The endogenous uncertainty is not a
problem here: given an investment plan we can immediately deduce the
survival probability for each link, which enables us to compute each
scenario probability and hence the path length expectations.

Unfortunately, this straightforward approach is impractical because we
must consider a billion scenarios to compute the fitness of each
chromosome.  In fact a major issue when solving stochastic problems by
metaheuristics is the fitness computation, and there are 3 common
approaches \cite{BiaEtc}:
\begin{itemize}
\item
Use a closed-form expression to compute exact fitness.
\item
Use a fast approximation to an expensive closed-form expression.
\item
Estimate fitness by sampling scenarios, as in the field of Simulation
Optimisation.
\end{itemize}
We propose an alternative approach: compute fitness exactly by
exploiting symmetries between scenarios, in order to bundle many of
them together so that they can be considered simultaneously.

\section{Exploiting symmetries between scenarios} \label{method}

\begin{figure}
\begin{center}
\includegraphics[scale=0.5]{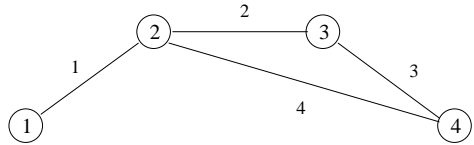}
\end{center}
\caption{A small network example}
\label{neteg}
\end{figure}

We shall illustrate our method using the simple example in Figure
\ref{neteg}.  The links $e=1 \ldots 4$ have lengths $t_e \equiv =1$,
$p_e \equiv 0.8$, $q_e \equiv 1$, $c_e \equiv 1$, $B=1$ and
$M_a=M_p=3.5$ so that both possible paths between nodes 1--4 are
allowed.  We must choose 1 link to invest in, to minimise the expected
shortest path length between nodes 1--4.  There are 16 scenarios, and
the optimal policy is to invest in link 1, giving an expected shortest
path length of 2.236.

\subsection{Merging scenarios}

Some scenarios can be considered together instead of separately.  For
example consider two scenarios 1001 (in which links 1 and 4 survive
but 2 and 3 do not) and 1101 (identical except that link 2 survives).
These scenarios have probabilities $0.8 \times 0.2 \times 0.2 \times
0.8=0.0256$ and $0.8 \times 0.8 \times 0.2 \times 0.8=0.1024$
respectively.  As link 3 does not survive, it is irrelevant whether or
not link 2 survives because it cannot be part of a shortest path (or
any path).  We can therefore merge these two scenarios into one, which
we shall write as 1i01 where ``i'' denotes {\it interchangeability\/}:
the values 0 and 1 for link 2 are interchangeable.  We shall refer to
a combined scenario such as 1i01 as a {\it multiscenario\/}.  As ``i''
includes both the 0 and 1 values, the probability associated with the
multiscenario is $0.8 \times 1.0 \times 0.2 \times 0.8=0.128$.

However, it is impractical to enumerate a billion scenarios then look
for ways of merging some of them.  Instead suppose we enumerate
scenarios by tree search on the random variables.  Consider a node in
the tree at which links $1 \ldots i$ have been realised, so that
random variables $s_1 \ldots s_{i-1}$ have been assigned values, and
we are about to assign a value to $s_i$ corresponding to link $e$.
Denote by $\ell_e$ the shortest source-sink path length including $e$,
under the assumption that all unrealised links survive; and denote by
$\ell_{\bar{e}}$ the shortest source-sink path length not including
$e$, under the assumption that all unrealised links fail (using $M_p$
when no path exists).  So $\ell_e$ is the minimum shortest path length
including $e$ in all scenarios below this policy tree node, while
$\ell_{\bar{e}}$ is the maximum shortest path length not including $e$
in the same scenarios.  They can easily be computed by temporarily
assigning $s_{i-1} \ldots s_n$ (where $|E|=n$) to 1 or 0 respectively,
and applying a shortest path algorithm.  Now if $\ell_e \ge
\ell_{\bar{e}}$ then the value assigned to $s_i$ is irrelevant: the
shortest path length in each scenario under this tree node is
independent of the value of $s_i$, so the values are interchangeable.

It is important to note that the order in which we assign the $s$
variables affects the size of the multiscenario set.  Three
multiscenario sets for the example are shown in Table \ref{multisets},
where $p$ is the multiscenario probability, 0 and 1 are the values of
the random variables corresponding to the links, and ``i'' denotes
interchangeable values: given the assignments to the left, the
objective value is independent of whether the random variable is set
to 0 or 1.  The set of size 7 corresponds to the permutation
corrsponding to the numerical link ordering 1234, the set of size 10
is the largest possible, and the set of size 5 is the smallest
possible.  Having derived the multiscenarios, we can replace the ``i''
entries by arbitrary values, for example 0.

\begin{table}
\begin{center}
\begin{tabular}{c@{\hspace{10mm}}c@{\hspace{10mm}}c}
\begin{tabular}{|rrrr|r|}
\hline
\multicolumn{4}{|c|}{links} & \\
3 & 2 & 4 & 1 & $p$\\
\hline
0 & i & 0 & i & 0.0400\\
0 & i & 1 & 0 & 0.0320\\
0 & i & 1 & 1 & 0.1280\\
1 & 0 & 0 & i & 0.0320\\
1 & 0 & 1 & 0 & 0.0256\\
1 & 0 & 1 & 1 & 0.1024\\
1 & 1 & 0 & 0 & 0.0256\\
1 & 1 & 0 & 1 & 0.1024\\
1 & 1 & 1 & 0 & 0.1024\\
1 & 1 & 1 & 1 & 0.4096\\
\hline
\end{tabular}
&
\begin{tabular}{|rrrr|r|}
\hline
\multicolumn{4}{|c|}{links} & \\
1 & 3 & 2 & 4 & $p$\\
\hline
0 & i & i & i & 0.2000\\
1 & 0 & i & 0 & 0.0320\\
1 & 0 & i & 1 & 0.1280\\
1 & 1 & 0 & 0 & 0.0256\\
1 & 1 & 0 & 1 & 0.1024\\
1 & 1 & 1 & 0 & 0.1024\\
1 & 1 & 1 & 1 & 0.4096\\
\hline
\end{tabular}
&
\begin{tabular}{|rrrr|r|}
\hline
\multicolumn{4}{|c|}{links} & \\
1 & 4 & 2 & 3 & $p$\\
\hline
0 & i & i & i & 0.2000\\
1 & 0 & 0 & i & 0.0320\\
1 & 0 & 1 & 0 & 0.0256\\
1 & 0 & 1 & 1 & 0.1024\\
1 & 1 & i & i & 0.6400\\
\hline
\end{tabular}
\end{tabular}
\end{center}
\caption{Three multiscenario sets for the small example}
\label{multisets}
\end{table}

\subsection{Stochastic dominance, symmetry and network reliability}

The above ideas have parallels in several literatures.  Firstly, one
way of viewing this form of reasoning is as {\it stochastic
  dominance\/} \cite{Lev}, a concept from the Decision Theory
literature: the objective function associated with one choice (0 or 1)
is at least as good as with another choice (1 or 0).  Because this
holds in every scenario, it is the simplest form of stochastic
dominance: {\it statewise\/} (or {\it zeroth order\/}) {\it
  dominance\/}.  However, this is usually defined as a strict
dominance by adding an extra condition: that one choice is strictly
better than the other in at least one state (or scenario).  In our
case neither value is better so this is a {\it weak dominance\/}.  In
fact we have two values that each weakly dominate the other, a
relationship that can be viewed as a {\it symmetry\/}: the tree is
exactly the same whichever value we use for a link.  But there does
not seem to be an accepted term such as ``stochastic symmetry'' for
this phenomenon.



However, in the Constraint Programming and Artificial Intelligence
literatures this type of symmetry is often used to reduce search tree
sizes (for non-stochastic problems): see Chapter 10 of \cite{RosEtc}
for a survey of such techniques.  Because the symmetry only occurs
under certain assignments to some other variables, it is a {\it
  conditional symmetry\/}, the condition being that certain other
assignments have occurred.  And because it is a symmetry on values in
the domain of a variable it is also a {\it value
  interchangeability\/}, a form of symmetry first investigated in
\cite{Fre} and since developed in many ways \cite{KarEtc}.  More
specifically, it is a form called {\it full dynamic
  interchangeability\/} \cite{KarEtc,Pre}, the word {\it dynamic\/}
having a similar meaning to {\it conditional\/} here.  Though
interchangeability has been the subject of considerable research, a
drawback is that it does not seem to occur in many real applications
\cite{ChoNou,Nea}; we believe that it will occur more often in
stochastic settings such as this one.

The Network Reliability literature describes methods for evaluating
and approximating the reliability of a network.  These include ways of
pruning irrelevant parts of a network that have connections to our
approach, though we have not found a direct parallel.  For a
discussion of these ideas see \cite{Col}.

\subsection{Application to the earthquake problem: single path}

For the earthquake problem we used a simple hill-climbing algorithm to
find a good permutation, based on 2- and 3-exchange moves, and
accepting moves that improve or leave unchanged the number of
multiscenarios.  The results are given in Table \ref{redtab} for each
source-sink pair considered separately, and took several minutes each
to compute.  The table shows the instances numbered 1--5, the source
and sink, the chosen constant $M_a$, the best link permutation found,
and the size of the corresponding multiscenario set.

\begin{table}
\begin{center}
\begin{tabular}{|r|r|rrrrrrrrrrrrrrr|r|}
\hline
instance & pair & \multicolumn{15}{c|}{link permutation} & multiscenarios\\
\hline
1 & 14--20 &
20 & 22 & 2 & 21 & 3 & 17 & 18 & 6 & 25 & 15 & 1 & 23 & 4 & 24 & 11 & 69\\
&& 14 & 26 & 30 & 13 & 19 & 12 & 10 & 29 & 28 & 7 & 16 & 27 & 5 & 9 & 8 &\\
\hline
2 & 14--7 &
20 & 16 & 10 & 25 & 17 & 14 & 1 & 28 & 9 & 2 & 13 & 3 & 8 & 11 & 4 & 45\\
&& 30 & 12 & 29 & 26 & 21 & 15 & 5 & 6 & 24 & 23 & 7 & 18 & 19 & 22 & 27 &\\
\hline
3 & 12--18 &
18 & 22 & 23 & 9 & 21 & 20 & 2 & 12 & 24 & 8 & 7 & 11 & 26 & 17 & 13 & 79\\
&& 3 & 6 & 27 & 14 & 16 & 10 & 28 & 1 & 19 & 15 & 29 & 4 & 30 & 5 & 25 &\\
\hline
4 & 9--7 &
1 & 3 & 28 & 29 & 10 & 13 & 20 & 18 & 8 & 11 & 9 & 27 & 12 & 16 & 17 & 26\\
&& 21 & 30 & 19 & 25 & 24 & 2 & 4 & 14 & 22 & 5 & 23 & 26 & 7 & 6 & 15 &\\
\hline
5 & 4--8 &
6 & 12 & 24 & 5 & 8 & 18 & 4 & 19 & 9 & 3 & 21 & 23 & 28 & 7 & 10 & 124\\
&& 29 & 11 & 20 & 13 & 30 & 17 & 2 & 26 & 16 & 22 & 15 & 27 & 14 & 1 & 25 &\\
\hline
\end{tabular}
\end{center}
\caption{Scenario reduction results for the earthquake problem}
\label{redtab}
\end{table}

Note that the choice of permutation has a significant effect.  For the
5 instances, if we use the numerical link ordering we obtain
multiscenario sets of sizes 4944, 4154, 5268, 87 and 1488
respectively, but we might be more unlucky: we sampled 10 random
permutations per instance and found worst-case multiscenario set sizes
31124, 115760, 21200, 994 and 7408 respectively.  These are much
better than 1 billion but considerably worse than the best sets we
found, showing the advantage of searching for a good permutation.


Table \ref{red97} shows in full the multiscenario set for the 4th
instance, which has the smallest set.  The table shows that the
expected shortest path length from 9--7 is independent of the survival
or failure of links 1, 2, 3, 4, 5, 8, 15, 18, 19, 20, 21, 22, 23, 24,
25, 26, 27, 28, 29 and 30; it depends only on the remaining 10 links
6, 7, 9, 10, 11, 12, 13, 14, 16 and 17.  Therefore the multiscenario
set for this instance should be no larger than $2^{10}=1024$.  But
among these 10 links there are interchangeabilities: for example if
link 10 fails then the expected length is independent of the survival
or failure of link 13, but if link 10 survives then what happens to
link 13 is important: if link 13 also survives (last line) then no
other link matters because the 9--7 shortest path is available.

\begin{table}
\begin{center}
\begin{tabular}{|rrrrrrrrrrrrrrrrrrrrrrrrrrrrrr|}
\hline
1&3&28&29&10&13&20&18&8&11&9&27&12&16&17&21&30&19&25&24&2&4&14&22&5&23&26&7&6&15\\
\hline
i&i&i&i&0&i&i&i&i&0&i&i&i&i&i&i&i&i&i&i&i&i&i&i&i&i&i&i&i&i\\
i&i&i&i&0&i&i&i&i&1&0&i&i&i&i&i&i&i&i&i&i&i&i&i&i&i&i&i&i&i\\
i&i&i&i&0&i&i&i&i&1&1&i&0&i&i&i&i&i&i&i&i&i&i&i&i&i&i&0&i&i\\
i&i&i&i&0&i&i&i&i&1&1&i&0&i&i&i&i&i&i&i&i&i&i&i&i&i&i&1&0&i\\
i&i&i&i&0&i&i&i&i&1&1&i&0&i&i&i&i&i&i&i&i&i&i&i&i&i&i&1&1&i\\
i&i&i&i&0&i&i&i&i&1&1&i&1&i&i&i&i&i&i&i&i&i&i&i&i&i&i&i&i&i\\
i&i&i&i&1&0&i&i&i&0&i&i&i&0&i&i&i&i&i&i&i&i&i&i&i&i&i&i&i&i\\
i&i&i&i&1&0&i&i&i&0&i&i&i&1&0&i&i&i&i&i&i&i&i&i&i&i&i&i&i&i\\
i&i&i&i&1&0&i&i&i&0&i&i&i&1&1&i&i&i&i&i&i&i&0&i&i&i&i&i&i&i\\
i&i&i&i&1&0&i&i&i&0&i&i&i&1&1&i&i&i&i&i&i&i&1&i&i&i&i&i&i&i\\
i&i&i&i&1&0&i&i&i&1&0&i&i&0&i&i&i&i&i&i&i&i&i&i&i&i&i&i&i&i\\
i&i&i&i&1&0&i&i&i&1&0&i&i&1&0&i&i&i&i&i&i&i&i&i&i&i&i&i&i&i\\
i&i&i&i&1&0&i&i&i&1&0&i&i&1&1&i&i&i&i&i&i&i&0&i&i&i&i&i&i&i\\
i&i&i&i&1&0&i&i&i&1&0&i&i&1&1&i&i&i&i&i&i&i&1&i&i&i&i&i&i&i\\
i&i&i&i&1&0&i&i&i&1&1&i&0&0&i&i&i&i&i&i&i&i&i&i&i&i&i&0&i&i\\
i&i&i&i&1&0&i&i&i&1&1&i&0&0&i&i&i&i&i&i&i&i&i&i&i&i&i&1&0&i\\
i&i&i&i&1&0&i&i&i&1&1&i&0&0&i&i&i&i&i&i&i&i&i&i&i&i&i&1&1&i\\
i&i&i&i&1&0&i&i&i&1&1&i&0&1&0&i&i&i&i&i&i&i&i&i&i&i&i&0&i&i\\
i&i&i&i&1&0&i&i&i&1&1&i&0&1&0&i&i&i&i&i&i&i&i&i&i&i&i&1&0&i\\
i&i&i&i&1&0&i&i&i&1&1&i&0&1&0&i&i&i&i&i&i&i&i&i&i&i&i&1&1&i\\
i&i&i&i&1&0&i&i&i&1&1&i&0&1&1&i&i&i&i&i&i&i&0&i&i&i&i&0&i&i\\
i&i&i&i&1&0&i&i&i&1&1&i&0&1&1&i&i&i&i&i&i&i&0&i&i&i&i&1&0&i\\
i&i&i&i&1&0&i&i&i&1&1&i&0&1&1&i&i&i&i&i&i&i&0&i&i&i&i&1&1&i\\
i&i&i&i&1&0&i&i&i&1&1&i&0&1&1&i&i&i&i&i&i&i&1&i&i&i&i&i&i&i\\
i&i&i&i&1&0&i&i&i&1&1&i&1&i&i&i&i&i&i&i&i&i&i&i&i&i&i&i&i&i\\
i&i&i&i&1&1&i&i&i&i&i&i&i&i&i&i&i&i&i&i&i&i&i&i&i&i&i&i&i&i\\
\hline
\end{tabular}
\end{center}
\caption{Multiscenario set for instance 4}
\label{red97}
\end{table}

\subsection{Extension to multiple paths}

We aim to minimise the expected weighted sum of shortest path lengths
$\ell_i$ between several source-sink pairs:
\[
\mbox{\bf Minimise}\;z=\mathbb{E}\left\{\sum_i w_i \ell_i\right\}
\]
for weights $w_i$.  Unfortunately, there is likely to be little
interchangeability in this problem, especially if (as we would expect)
the pairs are chosen to cover most of the network: for a given link to
be irrelevant to the lengths of several paths is much less likely than
for one path.  But we can avoid this drawback by rewriting the
objective function as:
\[
\mbox{\bf Minimise}\;z=\sum_i w_i \mathbb{E}\{\ell_i\}
\]
so that each path is treated separately, and can be evaluated using
its own link permutation.  If we do this using the permutations shown
in Table \ref{redtab} the total number of multiscenarios is 343, so to
evaluate an investment plan we need consider only this many
multiscenarios: a very tractable.  Note that we compute fitness
exactly, so the optimal investment plan is guaranteed to occur in the
search space of the genetic algorithm (though the algorithm is not
guaranteed to find it).


\subsection{Other risk measures}

Note that it is easy to change the objective function in a genetic
algorithm.  SP researchers have recently explored {\it risk-averse\/}
disaster planning including transportation networks \cite{LiuEtc}, and
we can use risk-averse objective functions such as {\it conditional
  value-at-risk\/} (CVaR) for a single source-sink pair.  For multiple
pairs we can take a weighted sum of CVaRs.  For our method to work
well, we must optimise some function of statistical parameters
computed on each pair separately.



\section{Conclusion} \label{conclusion}

In future work we shall implement metaheuristic and exact algorithms
to solve the reduced problem.  We shall also apply the reduction
technique to other problems.


\bibliographystyle{plain}

\begin{thebibliography}{10}

\bibitem{BiaEtc}
L. Bianchi, M. Dorigo, L. M. Gambardella, W. J. Gutjahr.
A Survey on Metaheuristics for Stochastic Combinatorial Optimization.
{\it Natural Computing\/} {\bf 8}(2):239--287, 2009.

\bibitem{ChoNou}
B. Y. Choueiry, G. Noubir.
On the Computation of Local Interchangeability in Discrete Constraint
Satisfaction Problems.  {\it AAAI/IAAI\/}, 1998, pp. 326--333.

\bibitem{Col}
C. J. Colbourn.
Concepts of Network Reliability.
Wiley Encyclopedia of Operations Research and Management Science,
John Wiley \& Sons, Inc., 2010.

\bibitem{ColMar}
M. Colvin, C. T. Maravelias.
A Stochastic Programming Approach for Clinical Trial Planning
in New Drug Development.
{\it Computers and Chemical Engineering\/} {\bf 32}(11):2626--2642, 2008.

\bibitem{CraEtc}
B. Craenen, A.E. Eiben, E. Marchiori.
How to Handle Constraints with Evolutionary Algorithms.
Practical Handbook of Genetic Algorithms.
L. Chambers (ed.), 2001, pp. 341--361.

\bibitem{Fre}
E. C. Freuder.
Eliminating Interchangeable Values in Constraint Satisfaction Problems.
{\it National Conference on Artificial Intelligence\/}, 1991, pp. 227--233.

\bibitem{GoeGro}
V. Goel, I. E. Grossmann.
A Class of Stochastic Programs with Decision Dependent Uncertainty.
{\it Mathematical Programming\/} {\bf 108}(2):355--394, 2006.

\bibitem{KarEtc}
S. Karakashian, R. Woodward, B. Y. Choueiry, S. D. Prestwich, E. C. Freuder.
A Partial Taxonomy of Substitutability and Interchangeability.
{\it 10th International Workshop on Symmetry in Constraint Satisfaction
Problems\/}, 2010.  (Journal paper in preparation.)

\bibitem{Lev}
H. Levy.
Stochastic Dominance and Expected Utility: Survey and Analysis.
{\it Management Science\/} {\bf 38}:555--593, 1992.

\bibitem{LiuEtc}
C. Liu, Y. Fan, F. Ord\'{o}\~{n}ez.
A Two-Stage Stochastic Programming Model for Transportation Network Protection.
{\it Computers \& Operations Research\/} {\bf 36}:1582--1590, 2009.

\bibitem{Nea}
N. Neagu.
Studying Interchangeability in Constraint Satisfaction Problems.
{\it 8th International Conference on Principles and Practice of
Constraint Programming, Lecture Notes in Computer Science\/} vol. 2470,
2002, pp. 787--788.

\bibitem{PeeEtc}
S. Peeta, F. S. Salman, D. Gunnec, K. Viswanath.
Pre-Disaster Investment Decisions for Strengthening a Highway Network.
{\it Computers \& Operations Research\/} {\bf 37}:1708--1719, 2010.

\bibitem{Pre}
S. D. Prestwich.
Full Dynamic Interchangeability with Forward Checking and Arc
Consistency.  {\it Workshop on Modeling and Solving Problems With
Constraints\/}, Valencia, 2004.

\bibitem{RosEtc}
F. Rossi, P. van Beek, T. Walsh (eds.).
Handbook of Constraint Programming.
Foundations of Artificial Intelligence series, Elsevier, 2006.

\bibitem{SolEtc}
S. Solak, J.-P. B. Clarke, E. L. Johnson, E. R. Barnes.
Optimization of R\&D Project Portfolios Under Endogenous Uncertainty.
{\it European Journal of Operational Research\/} {\bf 207}:420--433, 2010.

\end{thebibliography}

\end{document}